\documentclass[10pt,twocolumn,letterpaper]{article}

\usepackage[pagenumbers]{cvpr} 

\usepackage{graphicx}
\usepackage{amsmath}
\usepackage{amssymb}
\usepackage{booktabs}
\newcommand{\matr}[1]{\mathbf{#1}}

\usepackage[pagebackref,breaklinks,colorlinks]{hyperref}

\usepackage[capitalize]{cleveref}
\crefname{section}{Sec.}{Secs.}
\Crefname{section}{Section}{Sections}
\Crefname{table}{Table}{Tables}
\crefname{table}{Tab.}{Tabs.}

\def\cvprPaperID{5515} 
\def\confName{CVPR}
\def\confYear{2022}

\begin{document}

\title{CpT: Convolutional Point Transformer for 3D Point Cloud Processing}

\author{Chaitanya Kaul\\
University of Glasgow\\
Glasgow, UK\\
{\tt\small chaitanya.kaul@glasgow.ac.uk}
\and
Joshua Mitton\\
University of Glasgow\\
Glasgow, UK\\
{\tt\small j.mitton.1@research.gla.ac.uk}
\and
Hang Dai\\
MBZUAI\\
Abu Dhabi, UAE\\
{\tt\small hang.dai@mbzuai.ac.ae}
\and
Roderick Murray-Smith\\
University of Glasgow\\
Glasgow, UK\\
{\tt\small roderick.murray-smith@glasgow.ac.uk}
}
\maketitle

\begin{abstract}
   We present CpT: Convolutional point Transformer -- a novel deep learning architecture for dealing with the unstructured nature of 3D point cloud data. CpT is an improvement over existing attention-based Convolutions Neural Networks as well as previous 3D point cloud processing transformers. It achieves this feat due to its effectiveness in creating a novel and robust attention-based point set embedding through a convolutional projection layer crafted for processing dynamically local point set neighbourhoods. The resultant point set embedding is robust to the permutations of the input points. Our novel CpT block builds over local neighbourhoods of points obtained via a dynamic graph computation at each layer of the networks' structure. It is fully differentiable and can be stacked just like convolutional layers to learn global properties of the points. We evaluate our model on standard benchmark datasets such as ModelNet40, ShapeNet Part Segmentation, and the S3DIS 3D indoor scene semantic segmentation dataset to show that our model can serve as an effective backbone for various point cloud processing tasks when compared to the existing state-of-the-art approaches. 
\end{abstract}

\section{Introduction}
\label{sec:intro}
3D data takes many forms. Meshes, voxels, point clouds, multi view 2D images, RGB-D are all forms of 3D data representations. Amongst these, the simplest raw form that 3D data can exist in is a discretized representation of a continuous surface. This can be visualized as a set of points (in $\mathbb{R}^3$) sampled from a continuous surface. Adding point connectivity information to such 3D point cloud representations creates representations known as 3D meshes. Deep learning progress on processing 3D data was initially slow primarily due to the fact that early deep learning required a structured input data representation as a prerequisite. Thus, raw sensor data was first converted into grid-like representations such as voxels, multi view images, or data from RGB-D sensors was used to interpret geometric information. However, such data is computationally expensive to process, and in many cases, it is not the true representation of the 3D structure that may be required for solving the task. Modern applications of point clouds require a high amount of data processing. This makes searching for salient features in their representations a tedious task. Processing 3D meshes (along with added 3D point cloud information), requires dealing with their own sets of complexities and combinatorial irregularities, but such structures are generally efficient to process due to knowledge of point set connectivity. This motivation has sparked interest in processing 3D points directly at a point level instead of converting the points to intermediate representations. Lack of any general structure in the arrangement of points serves as a challenge in the ability to process them. This is due to the fact that point clouds are essentially set representations of a continuous surface in a 3D space. The seminal works on processing points proposed approaches to deal with this lack of order by constructing set based operations to ingest 3D points directly. They then created a symmetric mapping of such set based representations in a high dimensional space \cite{pointnet, deepsets}. This representation was further processed by symmetry preserving operations to create a feature representation of the point cloud, before passing them through multi layer perceptrons for solving the task. Representations created on a per-point basis are generally not robust as there is no concept of locality in the data representations used to create them. Various methods have been introduced in literature to add the notion of locality to point set processing \cite{pointnet++, pointcnn, spidercnn, dgcnn, sawnet, fatnet}.

Recently proposed transformer architectures have found great success in language tasks due to their ability to handle long term dependencies well \cite{transformer}. These models have been successfully applied to various computer vision applications \cite{detr, vit} and are already even replacing the highly successful CNNs as the de facto approach in many applications \cite{cvt, swintransformer}. Even though various domains of vision research have adopted the transformer architecture, their applications to 3D point cloud processing are still very limited \cite{pointtransformer, pct}. This leaves a gap in the current research on processing points with transformer based structures.

The basic intuition of our work lies in two simple points. First, transformers have been shown to be intrinsically invariant to the permutations in the input data \cite{transformersareequivariant} making them ideal for set processing tasks. Second, 3D points processed by most existing deep learning methods exploit local features to improve performance. However, these techniques treat points at a local scale to keep them invariant to input permutations leading to neglecting geometric relationships among point representations, and the inability of said models to capture global concepts well. To address these points, we propose {\it CpT: Convolutional point Transformer}. CpT differs from existing point cloud processing methods due to the following reasons,
\begin{itemize}
    \item CpT uses a $K$-nearest neighbour graph at every layer of the model. Such a dynamic graph computation \cite{dgcnn} requires a method to handle the input data, in order to create a data embedding that can be fed into a transformer layer. Towards this end, we propose a novel Point Embedding Module that first constructs a dynamic point cloud graph at every stage of the network, and creates a point embedding to feed into a transformer layer for its processing.
    \item Transformers employ a dot product attention mechanism to create contextual embeddings. Such an attention mechanism works to enhance the learning of features in the data. However, it has been shown that adding sample wise attention to data can help improve the performance of a transformer model even further \cite{saint, tabbie, mastransformer, axialattention}. To this end, we propose to add an InterPoint Attention Module to the Transformer which learns to enhance the output by learning to relate each point in the input to every other point. This helps capture better geometric relationships between the points and aids in better learning of the local and global concepts in the data. The resultant transformer block, i.e., the CpT layer (shown in Figure \ref{cpt_layer_diagram}) is a combination of a dot product attention operation, followed by an InterPoint Attention module. The $Q,K,V$ attention projections in this block are convolutional in nature to facilitate learning spatial context.
\end{itemize}

We evaluate CpT on multiple 3D point cloud datasets for classification, part segmentation, and semantic scene parsing tasks. We perform extensive experiments and multiple studies and ablations to show the robustness of our proposed model. Our results show that CpT can serve as an accurate and effective backbone for various point cloud processing tasks.   

The rest of the paper is organized as follows. Section \ref{sec:lit} gives an overview of the field of 3D point processing with deep learning and attention mechanisms used in vision applications which formed the basis of our study. We also review Transformer architectures proposed for vision applications. Section \ref{sec:cpt} gives a detailed explanation of CpT and its layers. We explain in detail the model structure and how the the proposed novel Point Embedding Module and the InterPoint Attention Module integrates into the setting. The evaluation of our work is presented in Section \ref{sec:expres} where we detail our experiments and results. Finally, Section \ref{sec:concfw} presents our conclusions and future directions of research. 

\section{Related Literature}
\label{sec:lit}

\textbf{Processing Point Clouds using Deep Learning.}
The lack of a grid-like structure to points, makes applying convolutions directly on them a tedious task. Due to this reason, all previous works in processing 3D points using deep learning, required the points to be first converted to a structured representation like voxels, depth maps, multiple 2D views of an object etc and then process the resultant representation with conventional deep models \cite{mvcnn, voxelnet, Voxnet}. This process has massive computational overheads which result from first converting the data to the grid like representation and then training large deep learning models on it. The first methods that proposed to treat points are set embeddings in a 3D space were PointNet \cite{pointnet} and Deep Sets \cite{deepsets}. They took approaches of creating data embeddings that preserve point set symmetry using permutation invariance and permutation equivariance respectively. This drastically reduced the computational overhead as the models used to train the data worked directly on the raw data points, and the models used to train on such data were not very large scale, and yet accurate. These models however, suffered from the drawback of only looking globally at the points. This meant that the higher dimension embeddings created by such methods were not robust to occlusions as the representations only took the particular input point into consideration while creating its representation. This drawback was tackled in works where local neighbourhoods of points were taken into consideration to compute per point representations. PointNet++ \cite{pointnet++} used farthest point sampling to estimate the locality of points first, before processing them via weight-shared multilayer perceptrons (MLPs). ECC Nets \cite{edgeconv} proposed the Edge Conv operation over a local neighbourhoods of points. Dynamic Graph CNNs \cite{dgcnn} created local neighbourhoods by computing a graphical representation of the points before every shared-weighted MLP layer to create a sense of locality in the points. Parameterized versions of convolution operations for points have also been proposed such as Spider CNN \cite{spidercnn}, and PointCNN \cite{pointcnn}. Previous work with kernel density functions to weight point neighbourhoods \cite{pointconv} treated convolutional kernels as non linear functions of the 3D points containing both weight and density estimations. The notion of looking both locally and globally at points was proposed in SAWNet \cite{sawnet}. Other notable approaches that process 3D points as a graph either densely connect local neighbourhoods \cite{pointweb}, operate on a superpoint graph to create contextual relationships \cite{SPG}, or use spectral graph convolutions to process the points \cite{pointsspecgraph}. \\

\textbf{Attention Mechanisms in Vision.} Various attention mechanisms exist in the deep learning literature, with one of the first works introducing them for natural language understanding being \cite{variousnlpattention}. The first attention mechanisms applied to vision were based on self attention via the squeeze-and-excite operation of SE Nets \cite{senet}. This attention mechanism provided channel-wise context for feature maps and was widely successful in increasing the accuracy over non attention based methods on ImageNet. It was extended in \cite{selfattention1} to non local networks, and in \cite{selfattention2} to fully convolutional networks for medical imaging applications. One of the first works that combined channel-wise self attention with spatial attention for images \cite{selfattncbam, selfattnfocusnet} first created attention maps across the entire feature map, followed by a feature map re-calibration step to only propagate the most important features in the networks forward. FatNet \cite{fatnet} successfully applied this concept to point clouds by first applying spatial attention over feature groups of local regions of points, followed by feature weighting using a squeeze-and-excite operation. 

\textbf{Transformers in 2D and 3D Vision.} A concurrent line of upcoming work applies the dot product attention to input data for its efficient processing. The Vision Transformer (ViT) \cite{vit} was the first real application of the Transformer to vision. It employed a Transformer encoder to extract features from image patches of size $16\times16$. An embedding layer converted the patches into a transformer-friendly representation and added positional embedding to them. This general structure formed the basis of initial transformer based research in vision, but required large amounts of data to train. DETR \cite{detr} was the first detection model created by processing the features of a Convolutional Neural Network by a transformer encoder, before adding a bipartite loss to the end to deal with a set based output. The Convolutional Vision Transformer (CvT) \cite{cvt} improved over the ViT by combining convolutions with Transformers via a novel convolutional token embedding and a convolutional projection layer. The recently proposed Compact Convolutional Transformer (CCT) \cite{cct} improve over the ViT and CvT models by showing that with the right size and tokenization, transformers can perform head-to-head with state of the art CNN models on small scale datasets. The Swin Transformer \cite{swintransformer} proposed a hierarchical transformer network whose representations are computed with shifted windows. 

The application of the transformer to point cloud processing is fairly limited. Existing works include the Point Transformer \cite{pointtransformer} which applies a single head of self attention on the points to create permutation invariant representations of local patches obtained via $K$-NN and furthest point sampling, and the Point Cloud Transformer \cite{pct} which applies a novel offset attention block to 3D point clouds. Over work extends over the existing literature to create a novel transformer block that performs feature wise as well as point wise dot product attention on the input for its accurate and effective processing.

\section{Convolutional Point Transformer}
\label{sec:cpt}

\begin{figure*}[htb]
\begin{center}
\includegraphics[width=1.0\linewidth]{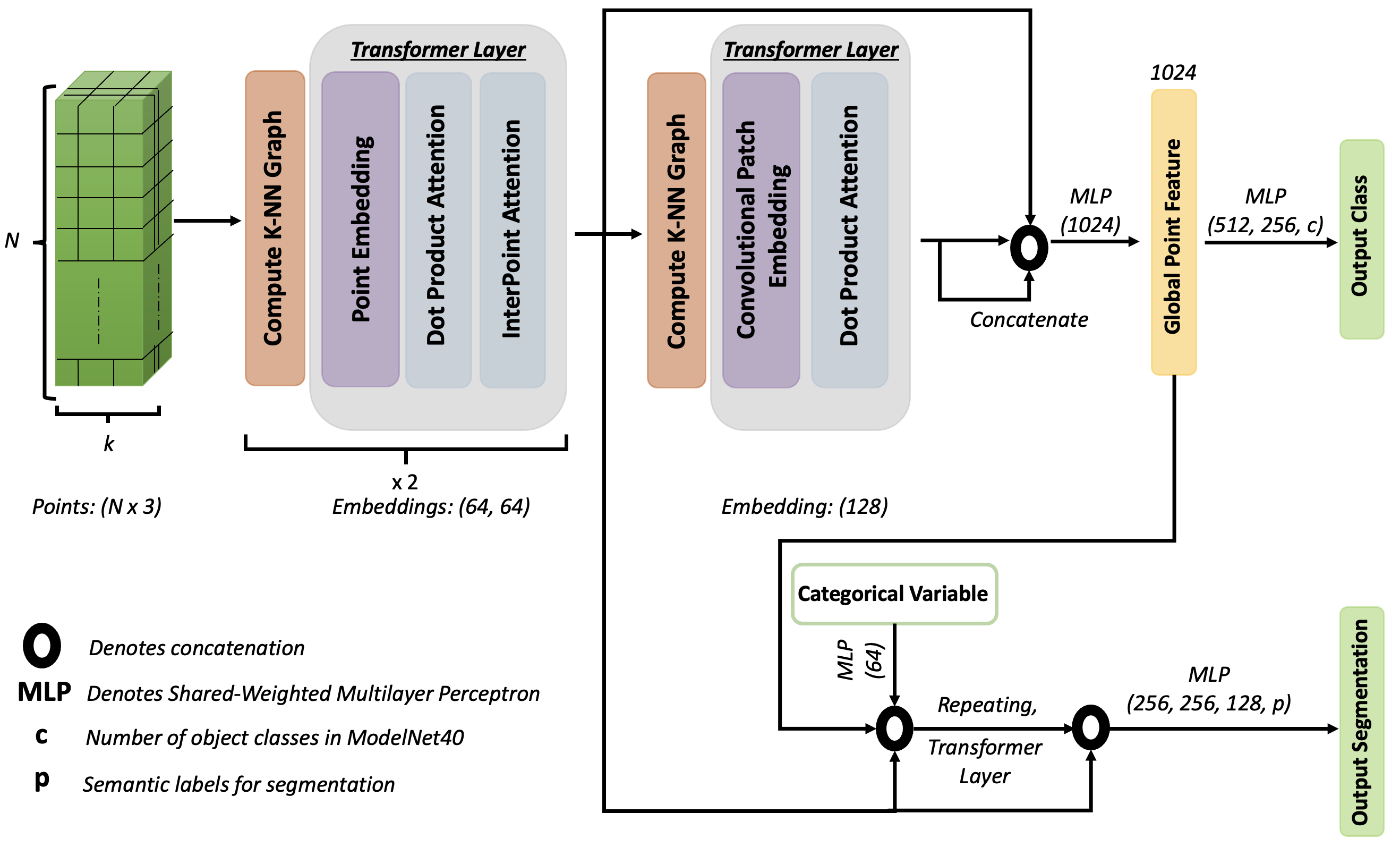}
\end{center}
\caption{The proposed CpT  architecture for the classification and segmentation tasks. Upper path: classification over $c$ classes. Lower path: segmentation over the $p$ semantic labels. The transformer layer in the segmentation branch does not contain InterPoint Attention. Flow of information is sequential from left to right.}
\label{cpt_Arch}
\end{figure*}

\textbf{Notation.} We denote a set of $N$-points with a $D$-dimensional embedding in the $l$th CpT layer as, $\matr{X}^l = \{ \mathbf{x}^l_1, \dots , \mathbf{x}^l_N \} $, where $\mathbf{x}^l_i \in \mathbb{R}^D$. For $K$ nearest neighbours of point $i$ ($i \in \{ 1 \dots N \}$), we define the set of all nearest neighbours for that point to be $\Delta\matr{X}^l_i = \{ \mathbf{x}^l_{j_1} - \mathbf{x}^l_{i}, \dots , \mathbf{x}^l_{j_K} - \mathbf{x}^l_{i} \} $, where $j_k: j_k \in \{ 1, \dots , N \} \wedge (j_k \ne i) $. 

The overall architecture of the Convolutional point Transformer (CpT) is shown in Figure \ref{cpt_Arch}. Our main contributions are the Point Embedding Module (Section \ref{pointembedding}) and the InterPoint Attention Module with a convolutional attention projection (Section \ref{cptlayer}). When an input point cloud of size $N \times 3$ is passed through the architecture, a graph of the points is computed via finding its $K$-Nearest Neighbours based on Euclidean distance. This representation is then passed through a point embedding layer that maps the input data into a representation implicitly inclusive of the nearest neighbours of the points. This is done via a 2D convolution operation whose degree of overlap across points can be controlled through the length of the stride. A dot product attention operation is then applied to this embedded representation which is followed by an InterPoint Attention Module. The dot product attention can be seen as learning relevant features of a points embedding as a function of it's K nearest neighbours. Such an attention mechanism learns to attend to the features of the points rather than the points themselves (column-wise matrix attention), i.e. for a set of points in a batch, it learns to weight individual feature transformations. The InterPoint Attention on the other hand can be interpreted as learning the relationships between different the points themselves, within a batch (a row-wise matrix attention operating per point embedding, rather than per individual feature of the points). This forms one layer of the CpT. We update the graph following the InterPoint Attention operation which is then passed into the next CpT layer. A third and final CpT layer with a higher feature space embedding dimension is used without InterPoint Attention to learn the representation of the point cloud. InterPoint attention in the deeper layers does not provide a lot of context, as transformers eventually can learn relations that cross beyond the locality of their point set embedding dimensions. This also means that transformers are able to visualize inputs beyond their limited receptive field in the deeper layers, making them a better choice for feature space embedding compared to shared-weighted MLPs (1D Convolutions). Adding this attention block to the deeper layers only marginally improves performance but significantly increases computation. The output of each transformer layer is then concatenated and passed through a final shared-weighted MLP. This is then Global Max-Pooled to get a global feature vector which forms the representation of the point cloud. This representation is then processed further through a series of MLPs to classify the point cloud, or obtain a semantic point label for each point in the point cloud. Residual connections, layer normalization, and addition and MLP layers are used after the attention layers as in conventional transformer model structures. 

\subsection{Point Embedding Module}
\label{pointembedding}

The point embedding module takes a k-NN graph as an input. The general structure of the input to this module is denoted by $(B, f, N, \Delta\matr{X}^l_i) \in \mathbb{R}^{(B \times f \times N \times \Delta\matr{X}^l_i)}$, where $B$ and $f$ are the batch size and input features respectively. This layer is essentially a mapping function $F_\theta$ that maps the input into an embedding $E_\theta$. This operation is denoted by, $$\matr{y} = \matr{F_\theta}(I(B, f, N, \Delta\matr{X}^l_i)), \matr{y} \in \mathbb{R}^{(B \times f \times \matr{E_\theta})}. $$

There are many choices available for the mapping function $\matr{F_\theta}$. We use a convolution operation with a fixed size padding and stride. This allows us to train this module in a end-to-end setting along with the rest of the network as it is fully differentiable and can be plugged anywhere in the architecture.

\subsection{Convolutional point Transformer Layer}
\label{cptlayer}

\begin{figure}[htbp]
\begin{center}
\includegraphics[width=1.0\linewidth]{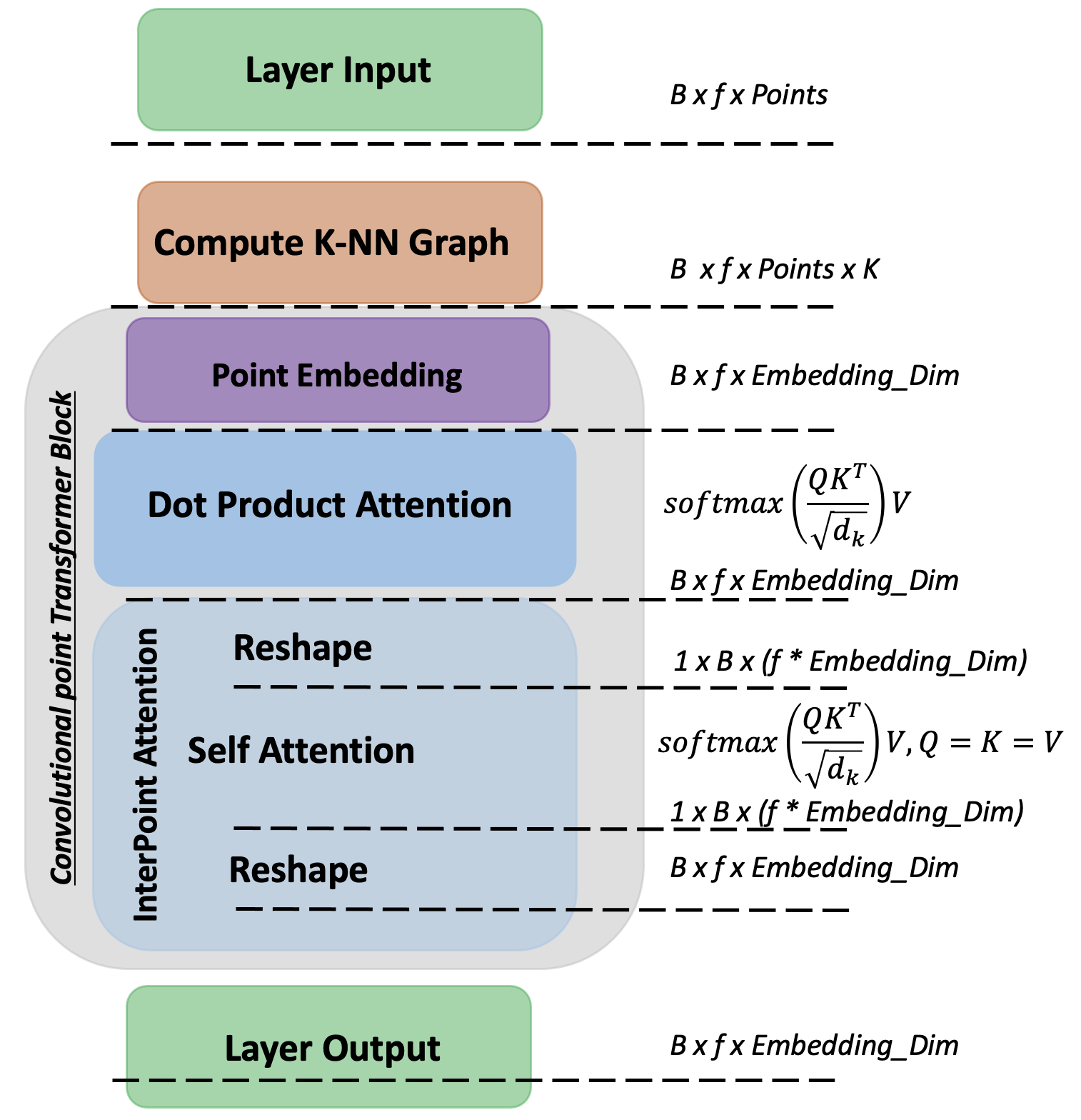}
\end{center}
\caption{The CpT Transformer Block with Dot Product Attention and InterPoint Attention. Flow of information is sequential from top to bottom.}
\label{cpt_layer_diagram}
\end{figure}

Our Convolutional point Transformer uses a combination of dot product and self attention to propagate the most important, salient features of the input through the network. The CpT layer leverages spatial context and moves beyond a fully connected projection, by using a convolution layer to sample the attention matrices. Instead of creating a more complicated network design, we leverage the ability of convolutions to learn relevant feature sets for 3D points. Hence, we replace the original fully connected layers in the attention block with depthwise convolution layers which forms our convolutional projection to obtain the attention matrices. The general structure of the CpT layer is shown in Figure \ref{cpt_layer_diagram}. The functionality of the rest of the transformer block is similar to ViT \cite{vit} where normalization and feedforward layers are added after every attention block and residual mappings are used to enhance the feature learning. \\

Formally, the projection operation of the CpT layer is denoted by the following operation, $$z^{q/k/v}_i = Convolution(z_i, p, s), $$ where $z^{q/k/v}$ is the input for the $Q/K/V$ attention matrices for the $i$-th layer and $z_i$ is the input to the convolution block. The convolution operation is implemented as a depthwise separable convolution operation with tuples of kernel size $p$ and stride $s$. We now formalize the flow of data through the entire CpT block for a batch size $B$ as, $$out^a_i = \beta(DA(z^{q/k/v}_i)) + z^{q/k/v}_i, $$ where $out^a_i$ is the output of the dot product attention block, $DA$ is the dot product attention given by $DA(\cdot) = softmax(\frac{QK^T}{\sqrt{d}}V)$. $\sqrt{d}$ is used to scale the attention weights to avoid gradient instabilities. $\beta$ is the layer normalization operation. The addition denotes a residual connection. This output is further processed as in a vanilla transformer in the following way, $$out^b_i = \beta(FF(out^a_i)) + out^a_i$$ FF here denotes the feedforward layers. The InterPoint attention is then computed on this output as, $$ out^c_i = \beta(DA(out^b_i)^B_{i=1}) + out^b_i $$ which is then processed in a similar manner by layer normalization and feedforward layers as, $$ out^d_i = \beta(FF(out^b_i) + out^b_i $$ Following this step, the graph is recomputed and the process restarts for the $l+1$-th layer.

\section{Experiments and Results}
\label{sec:expres}

We evaluate our model on three different datasets for point cloud classification, part segmentation and semantic scene segmentation. For classification, we use the benchmark ModelNet40 dataset \cite{modelnet}, for object part segmentation \cite{shapenet}, we use the ShapeNet Part dataset. We use the Stanford Large-Scale 3D Indoor Spaces (S3DIS) dataset \cite{s3dis} for semantic scene segmentation.

\subsection{Implementation Details}
\label{implementationdet}
Unless stated otherwise, all our models are trained in PyTorch on a batch size of 32 for 250 epochs. SGD with an initial learning rate of 0.1 and momentum 0.9 is used. We use a cosine annealing based learning rate scheduler. The momentum for batch normalization is 0.9. Batch Normalization decay is not used. Dropout, wherever used, is used with a rate of 0.5. Custom learning rate schedules are used for the segmentation tasks after initial experimentation. For the classification and 3D indoor scene segmentation tasks, we compute dynamic graphs using 20 nearest neighbours, while for the part segmentation task, we use 40 nearest neighbours. We use a NVIDIA Titan RTX GPU for our experiments.

\subsection{Classification with ModelNet40}

The ModelNet40 dataset \cite{modelnet} contains meshes of 3D CAD models. A total of 12,311 models are available belonging to 40 categories, split into a training-test set of 9,843-2,468 respectively. We use the official splits provided for all our experiments and datasets to keep a fair comparison. In terms of data pre-processing, we follow the same steps as \cite{pointnet}. We uniformly sample 1024 points from the mesh surface and recale the point cloud to fit a unit sphere. Data augmentation is used during the training process. We perturb the points with random jitter and scalings during the augmentation process. 

\begin{table}[h]
\begin{center}
\begin{tabular}{p{3.2cm} p{1.8cm} p{1.8cm}}
\hline
Method & Class Accuracy (\%) & Instance Accuracy (\%)\\
\hline\hline
3D ShapeNets \cite{shapenet} & 77.3 & 84.7\\
VoxNet \cite{Voxnet} & 83.0 & 85.9\\
PointNet \cite{pointnet} & 86.0 & 89.2\\
PointNet++ \cite{pointnet++} & - & 90.7\\
SpiderCNN \cite{spidercnn} & - & 90.5\\
PointWeb \cite{pointweb} & 89.4 & 92.3 \\
PointCNN \cite{pointcnn} & 88.1 & 92.2 \\
Point2Sequence \cite{pointtosequence} & 90.4 & 92.6 \\
ECC \cite{edgeconv} & 83.2 & 87.4\\
DGCNN \cite{dgcnn} & 90.2 & 92.2\\
FatNet \cite{fatnet} & 90.6 & 93.2\\
KPConv \cite{kpconv} & - & 92.9 \\
SetTransformer \cite{settransformer} & & 90.4 \\
PCT \cite{pct} & - & 93.2 \\
PointTransformer \cite{pointtransformer} & \textbf{90.6} & 93.7 \\
\hline
CpT (Static Graph) & 90.2 & 91.9 \\
CpT & \textbf{90.6} & \textbf{93.9} \\
\hline
\end{tabular}
\end{center}
\caption{Classification results on the ModelNet40 dataset.}
\label{ModelNet40}
\end{table}

It can be seen from the results that CpT outperforms existing classification methods well, and is even marginally better than the Point Transformer \cite{pointtransformer}. As we take a graph computation approach in our model, we compare with other methods that process static graphs \cite{edgeconv} as well as dynamic graphs \cite{dgcnn} for point cloud classification. Here, even when we do not recompute the graph at every layer, CpT outperforms Dynamic Edge Conditioned Filters (ECC) \cite{edgeconv} and performs at par with DGCNN. Dynamic graph computations before every layer help CpT surpass the accuracy of all existing graph and non graph based approaches in literature, including marginally outperforming existing transformer based approaches \cite{pct, pointtransformer}. 

\subsubsection{Ablation Study}

In this section, we detail experiments of our extensive studies into the CpT architecture. We conducted a series of ablation studies to highlight how the different building blocks of CpT come together.\\

\textbf{Static v/s Dynamic Graphs.} We trained two CpT models for these experiments. The underlying structure of the models is the same as Figure \ref{cpt_Arch}, with the only difference being that the K-NN graph computation is only done before the first CpT Layer in the Static CpT model. The results for this experiment are shown in Table \ref{ModelNet40}. Computing the graph before each transformer layer helps boost CpT's instance accuracy by 2\% and achieves state-of-the-art results. \\

\textbf{Global Representations vs Graph Representations.} We compared dynamic graph computation with feeding point clouds directly into the CpT model. This equates to removing the K nearest neighbour step from the model in Figure \ref{cpt_Arch}. As the points were directly fed into CpT, we also replaced the Point Embedding Module with a direct parameterized relational embedding This relation was learnt using a convolution operation. The rest of the architecture remained unchanged. The resultant model trained on global representations performed surprisingly well. The results are summarized in Table \ref{globalvsgraph}. PointNet \cite{pointnet} is added to the table for reference as it also takes a global point cloud representation as an input. The CpT with dynamic graph computation outperforms both methods, but it is interesting to note that the CpT model that works directly on the entire point cloud manages a higher class accuracy than PointNet. 

\begin{table}[ht]
\begin{center}
\begin{tabular}{c | c}
\hline
Model & Class Accuracy (\%) \\
\hline\hline
PointNet \cite{pointnet} & 86.2 \\
CpT (No locality) & 87.6 \\
CpT (Dynamic Graph) & 90.6 \\
\hline
\end{tabular}
\end{center}
\caption{Comparison of different input representations.}
\label{globalvsgraph}
\end{table}

\textbf{Number of Nearest Neighbours for dynamic graph computation.} We also experiment with the number of nearest neighbours used to construct the graph. The rest of the architecture is the same as Figure \ref{cpt_Arch}.

\begin{table}[ht]
\begin{center}
\begin{tabular}{c | c}
\hline
k & Class Accuracy (\%) \\
\hline\hline
10 & 89.6 \\
20 & 90.6 \\
30 & 90.3 \\
40 & 90.4 \\
\hline
\end{tabular}
\end{center}
\caption{CpT with different number of nearest neighbours.}
\label{nnpoints}
\end{table}

CpT layers implemented with 20 nearest neighbour graphs performs the best. This is due to the fact that for large distances beyond a particular threshold, the euclidean distance starts to fail to approximate the geodesic distance. This leads to capturing points that may not lie in the true neighbourhood of the point while estimating its local representation. \\

\textbf{Effect of Random Point Dropout.} We test the robustness of CpT towards classifying sparse datasets by sampling point clouds at various resolutions from the ModelNet40 dataset and evaluating CpT's performance on them. We sample 512, 128 and 64 points for each CAD model in the dataset, keeping the input point clouds' resolution small. The results are summarized in Table \ref{pointdrop}.

\begin{table}[ht]
\begin{center}
\begin{tabular}{c | c}
\hline
Method & Class Accuracy (\%) \\
\hline\hline
CpT (512 points) & 88.4 \\
CpT (128 points) & 85.1 \\
CpT (64 points) & 83.2 \\
\hline
\end{tabular}
\end{center}
\caption{Evaluating the effect of random point dropout on CpT.}
\label{pointdrop}
\end{table}

CpT is robust to random point dropout. This is largely due to the fact that CpT is not tied to observing input data through receptive fields like CNNs in order to construct its feature space. CpT looks beyond local neighbourhoods due to its InterPoint attention module which helps it build context for the 3D points beyond the nearest neighbours of the graph. The constructed feature space is hence, a combination of both local as well as global 3D shape properties. 

\subsection{Segmentation results with ShapeNet Part}

\begin{figure*}[ht]
\begin{center}
\includegraphics[width=1.0\linewidth]{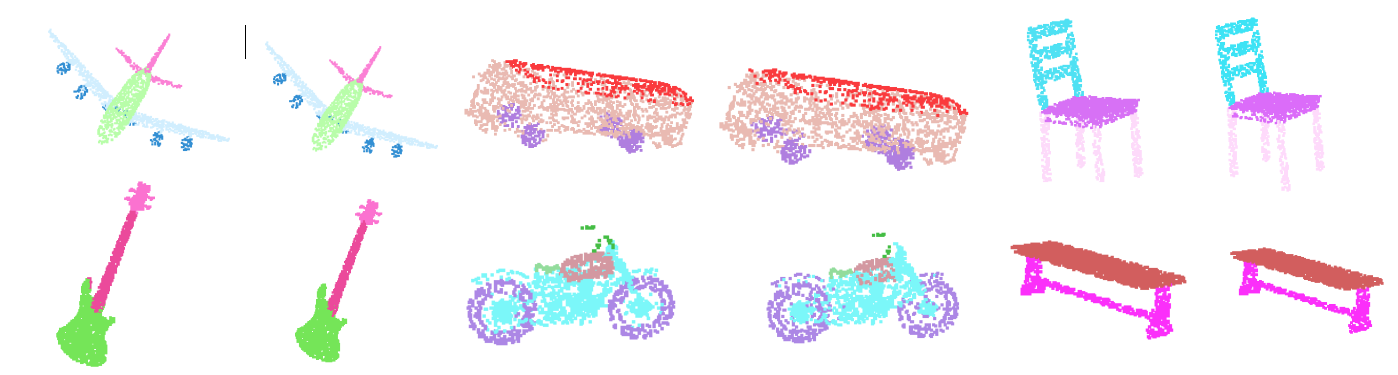}
\end{center}
\caption{Visualizing the segmentation results from the ShapeNet Part Dataset. From left to right, the images are alternatively the ground truth and the output prediction by CpT.}
\label{ShapeNetPart_Results}
\end{figure*}

The ShapeNet Part dataset \cite{shapenet} contains 16,881 3D shapes from 16 object categories. A total of 50 object parts are available to segment. We sample 2048 points from each shape and follow the official training-testing splits for our experiments. Our results on the dataset are summarized in Table \ref{part_seg}, while the visualizations produced by our model are shown in Figure \ref{ShapeNetPart_Results}. We train two CpT models for this task - the regular 3 layer CpT model and a larger 5 layer model, both with a batch size of 16 due to memory constraints. We see the the more compact model by itself outperforms most CNN based approaches for point cloud processing with KPConv \cite{kpconv} being the only exception. The larger 5 layer model outperforms PCT by 0.2\% mIoU while performs at par with the current state-of-the-art Point Transformer \cite{pointtransformer} model which is much larger in size than CpT (\# params).

\begin{table}[h]
\begin{center}
\begin{tabular}{p{3.2cm} | p{1.5cm} }
\hline
Method & IoU \\
\hline\hline
Kd-Net \cite{kdnet} & 82.3  \\
SO-Net \cite{sonet} & 84.6  \\
PointNet++ \cite{pointnet++} & 85.1 \\
SpiderCNN \cite{spidercnn} & 85.3 \\
SPLATNet \cite{splatnet} & 85.4 \\
PointCNN \cite{pointcnn} & 86.1 \\
PointNet \cite{pointnet} & 83.7 \\
DGCNN \cite{dgcnn} & 85.1 \\
FatNet \cite{fatnet} & 85.5 \\
PointASNL \cite{pointasnl} & 86.1 \\
RSCNN \cite{rscnn} & 86.2 \\
KPConv \cite{kpconv} & 86.4 \\
PCT \cite{pct} & 86.4 \\
PointTransformer \cite{pointtransformer} & \textbf{86.6} \\
\hline
CpT & 86.3 \\
CpT (5 layers) & \textbf{86.6} \\
\hline
\end{tabular}
\end{center}
\caption{Results on the ShapeNet Part Segmentation dataset.}
\label{part_seg}
\end{table}

\subsection{3D Semantic Segmentation with S3DIS}

\begin{figure*}[ht]
\begin{center}
\includegraphics[width=1.0\linewidth]{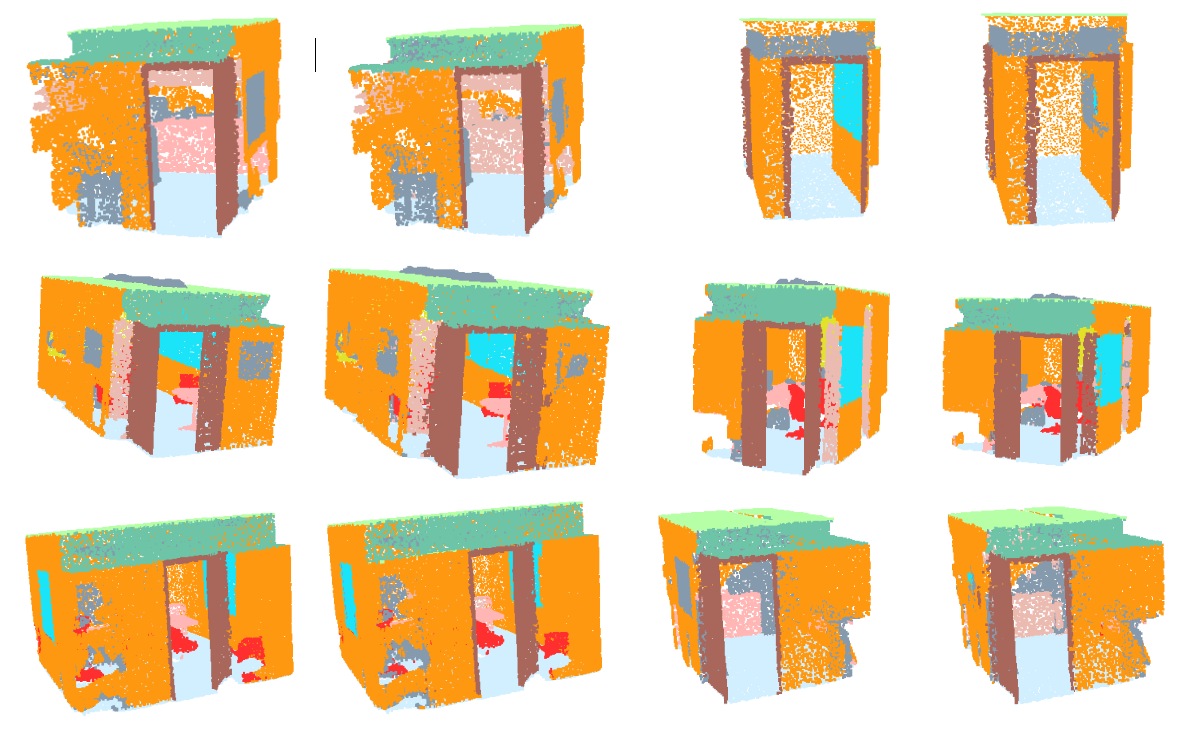}
\end{center}
\caption{Visualizing the segmentation results from the S3DIS 3D Indoor Scene Segmentation Dataset (Area 6). From left to right, the images are alternatively the ground truth and the output prediction by CpT.}
\label{S3DIS_Results}
\end{figure*}

The Stanford 3D Indoor Scene Segmentation dataset \cite{s3dis} contains 6 indoor scenes with a total of 272 rooms, making it a challenging dataset. Each point in the scene can belong to one of 13 object categories. We split the rooms are split into blocks where each block is represented by a 9-Dimensional vector (XYZ, RGB, normalized co-ordinates). 4096 points are sampled from each scene to form the inputs for the model. We train models in a 5-fold cross validation setting over the areas and report the average intersection over union (mIoU) results and the overall accuracy of the model for Area 6. Our results are summarized in Table \ref{sem_seg} and the visualizations produced by our model are illustrated in Figure \ref{S3DIS_Results}. We train PointNet, PointNet++, DGCNN and CpT on the S3DIS dataset with the same hyperparameters as described in Section \ref{implementationdet} with three exceptions. All models are trained for 150 epochs, the batch size for these experiments is set to 8, and the learning rate for the PointNet and PointNet++ models is set to 0.001 initially. CpT outperforms the methods compared with in terms of the mean intersection over union, showing its superior performance. The Point Transformer requires 4 Titan RTX GPUs to train on the S3DIS dataset and is hence not used in the comparison due to compute constraints.

\begin{table}[h]
\begin{center}
\begin{tabular}{p{3.2cm} | p{1.5cm}  p{1.5cm} }
\hline
Method & mIoU & mAcc \\
\hline\hline
PointNet \cite{pointnet} & 40.2 & 46.4 \\
PointNet++ \cite{pointnet++} & 51.7 & 66.1 \\
DGCNN \cite{dgcnn} & 58.4 & \textbf{77.2} \\
PCT \cite{pct} & 60.9 & 64.1 \\
\hline
CpT & \textbf{62.3} & 72.6 \\
\hline
\end{tabular}
\end{center}
\caption{Results on the S3DIS 3D indoor scene Segmentation dataset. Results are presented on Area 6.}
\label{sem_seg}
\end{table}

\section{Conclusions and Future Work}
\label{sec:concfw}

In this paper, we proposed the CpT: Convolutional point Transformer. We showed how transformers can be effectively used to process 3D points with the help of dynamic graph computations at each intermediate network layer. The main contributions of our work include, first, a Point Embedding Module capable of taking a dynamic graph as an input and transforming it into a transformer friendly data representation. Second, the InterPoint Attention Module which uses self attention to facilitate cross talk between the points in an arbitrary batch. Through this work, we have shown for the first time that dot product and self attention methods can be efficiently used inside a single transformer's layer for 3D point cloud processing (\cite{pointtransformer} only uses self attention while \cite{pct} only uses an offset attention operator). CpT outperforms most existing convolutional approaches for point cloud processing and performs competitively with the existing transformer based approaches on a variety of benchmark tasks. To improve performance, future directions of this research lie in learning to sample points uniformly along the manifold of the 3D point cloud to preserve its local shape. This can lead to learning better local representations of the data and in turn, creating models with improved accuracy. Our results already show that CpT is capable of taking local context and processing it effectively with global information present in the points. Further, to improve the performance of CpT points in each CpT layer can be processed at different resolutions (like in \cite{pointtransformer}). We believe that CpT can serve as an effective backbone for future point cloud processing tasks and be extended to various applications.

\section{Acknowledgements}

Chaitanya Kaul and Roderick Murray-Smith acknowledge funding from the QuantIC project funded by the EPSRC Quantum Technology Programme (grant EP/MO1326X/1) and the iCAIRD project, funded by Innovate UK (project number 104690). Joshua Mitton is supported by a University of Glasgow Lord Kelvin Adam Smith Studentship.

{\small
\bibliographystyle{ieee_fullname}
\bibliography{egbib}
}

\end{document}